%% file: main_arxiv.tex
\documentclass[10pt,twocolumn,letterpaper]{article}

\usepackage{cvpr}              %

\usepackage{graphicx}
\usepackage{amsmath}
\usepackage{amssymb}
\usepackage{booktabs}

\usepackage[pagebackref,breaklinks,colorlinks, allcolors=blue]{hyperref}

\usepackage[capitalize]{cleveref}
\crefname{section}{Sec.}{Secs.}
\Crefname{section}{Section}{Sections}
\Crefname{table}{Table}{Tables}
\crefname{table}{Tab.}{Tabs.}

\usepackage[accsupp]{axessibility}
\usepackage{bbding}
\newcommand\blfootnote[1]{%
  \begingroup
  \renewcommand\thefootnote{}\footnote{#1}%
  \addtocounter{footnote}{-1}%
  \endgroup
}

\input{math_commands.tex}

\input{macros}

\def\cvprPaperID{1411} %
\def\confName{CVPR}
\def\confYear{2022}

\begin{document}

\title{Balanced MSE for Imbalanced Visual Regression}

\author{Jiawei Ren\Mark{1},
Mingyuan Zhang\Mark{1},
Cunjun Yu\Mark{2},
Ziwei Liu\Mark{1} \Mark{\Envelope}
\\
\Mark{1}S-Lab, Nanyang Technological University
\\\Mark{2}School of Computing, National University of Singapore
\\
{\tt\small \{jiawei011, mingyuan001\}@e.ntu.edu.sg, cunjun.yu@comp.nus.edu.sg, ziwei.liu@ntu.edu.sg}
}
\maketitle

\blfootnote{\Mark{\Envelope} Corresponding author.}
\input{sections/0_abstract}
\input{sections/1_introduction}

\input{sections/2_related_works}
\input{sections/3_method}

\input{sections/4_experiments}
\input{sections/5_conclusion}

\input{sections/6_ack}
\clearpage
\appendix
\section*{Appendix}
\input{sections/appendix}

\clearpage
{\small
\bibliographystyle{ieee_fullname}
\bibliography{references}
}

\end{document}

%% file: math_commands.tex
\usepackage{amsmath,amsfonts,bm}

\def\eqref#1{equation~\ref{#1}}

\def\1{\bm{1}}

\DeclareMathAlphabet{\mathsfit}{\encodingdefault}{\sfdefault}{m}{sl}
\SetMathAlphabet{\mathsfit}{bold}{\encodingdefault}{\sfdefault}{bx}{n}

%% file: macros.tex
\usepackage{wrapfig}
\usepackage{makecell}
\usepackage{bm}

\numberwithin{equation}{section}

\newcommand\Mark[1]{\textsuperscript#1}

\def\equationautorefname~#1\null{Eq.~#1\null}
\def\figureautorefname~#1\null{Fig.~#1\null}
\def\tableautorefname~#1\null{Tab.~#1\null}
\def\sectionautorefname~#1\null{Sect.~#1\null}
\def\subsectionautorefname~#1\null{Sect.~#1\null}
\def\subsubsectionautorefname~#1\null{Sect.~#1\null}

\newcommand{\threesubsection}[1]{\textbf{#1}.~}

\def\shownotes{1} 
 \ifnum\shownotes=1
\newcommand{\authnote}[2]{{[#1: #2]}}
\else 
\newcommand{\authnote}[2]{{}}
\fi

\newcommand{\ev}{\mathbb{E}}
\newcommand{\rspace}{\mathbb{R}}
\newcommand{\gau}{\mathcal{N}}
\newcommand{\dy}{d\boldsymbol{y}}
\newcommand{\dyprime}{d\boldsymbol{y}^\prime}

\newcommand{\norm}[1]{\left\lVert#1\right\rVert}

\newtheorem{theorem}{Theorem}
\newtheorem{definition}{Definition}[section]

\newcommand{\pxy}{p(\boldsymbol{x}|\boldsymbol{y})}

\newcommand{\pyxtheta}{p(\boldsymbol{y}|\boldsymbol{x};\boldsymbol{\theta})}
\newcommand{\pyxtraintheta}{p_{\textup{train}}(\boldsymbol{y}|\boldsymbol{x};\boldsymbol{\theta})}
\newcommand{\pyxtrainthetauni}{p_{\textup{train}}(y|\boldsymbol{x};\boldsymbol{\theta})}
\newcommand{\pyxbaltheta}{p_{\textup{bal}}(\boldsymbol{y}|\boldsymbol{x};\boldsymbol{\theta})}
\newcommand{\pyxbalthetauni}{p_{\textup{bal}}(y|\boldsymbol{x};\boldsymbol{\theta})}

\newcommand{\pytrain}{p_{\textup{train}}(\boldsymbol{y})}
\newcommand{\pytrainuni}{p_{\textup{train}}({y})}

\newcommand{\pxtrain}{p_{\textup{train}}(\boldsymbol{x})}
\newcommand{\pyxtrain}{p_{\textup{train}}(\boldsymbol{y}|\boldsymbol{x})}
\newcommand{\pyxtrainuni}{p_{\textup{train}}({y}|\boldsymbol{x})}

\newcommand{\pybal}{p_{\textup{bal}}(\boldsymbol{y})}

\newcommand{\pxbal}{p_{\textup{bal}}(\boldsymbol{x})}
\newcommand{\pyxbal}{p_{\textup{bal}}(\boldsymbol{y}|\boldsymbol{x})}
\newcommand{\pyxbaluni}{p_{\textup{bal}}({y}|\boldsymbol{x})}

\newcommand{\yprime}{\boldsymbol{y}^\prime}
\newcommand{\pyprimetrain}{p_{\textup{train}}(\yprime)}
\newcommand{\pyprimetrainuni}{p_{\textup{train}}(y^\prime)}
\newcommand{\pyprimextrain}{p_{\textup{train}}(\yprime|\boldsymbol{x})}

\newcommand{\pyprimebal}{p_{\textup{bal}}(\yprime)}
\newcommand{\pyprimexbal}{p_{\textup{bal}}(\yprime|\boldsymbol{x})}
\newcommand{\pyprimexbaluni}{p_{\textup{bal}}(y^\prime|\boldsymbol{x})}
\newcommand{\pyprimexbaltheta}{p_{\textup{bal}}(\yprime|\boldsymbol{x};\boldsymbol{\theta})}

\newcommand{\ytarget}{\boldsymbol{y}_\textup{target}}

\newcommand{\ypred}{\boldsymbol{y}_\textup{pred}}
\newcommand{\sigmapred}{\sigma_\textup{noise}}

\newcommand{\Sigmapred}{\sigmapred^2 \boldsymbol{\textup{I}}}

\newcommand{\pysampletrainwoeq}{p_{\textup{train}}(\boldsymbol{y}_{(i)})}

\newcommand{\name}{Balanced MSE}
\newcommand{\benchmark}{IHMR}

%% file: sections/0_abstract.tex
\begin{abstract}

Data imbalance exists ubiquitously in real-world visual regressions, \emph{e.g.,} age estimation and pose estimation, hurting the model's generalizability and fairness. Thus, imbalanced regression gains increasing research attention recently. Compared to imbalanced classification, imbalanced regression focuses on continuous labels, which can be boundless and high-dimensional and hence more challenging.
In this work, we identify that the widely used Mean Square Error (MSE) loss function can be ineffective in imbalanced regression. We revisit MSE from a statistical view and propose a novel loss function, Balanced MSE, to accommodate the imbalanced training label distribution.
We further design multiple implementations of Balanced MSE to tackle different real-world scenarios, particularly including the one that requires no prior knowledge about the training label distribution. Moreover, to the best of our knowledge, Balanced MSE is the first general solution to high-dimensional imbalanced regression in modern context. Extensive experiments on both synthetic and three real-world benchmarks demonstrate the effectiveness of Balanced MSE. Code and models are available at \href{https://github.com/jiawei-ren/BalancedMSE}{github.com/jiawei-ren/BalancedMSE}.
\end{abstract}

%% file: sections/1_introduction.tex
\section{Introduction}

\begin{figure}[!t]
\centering
	\includegraphics[width=0.475\textwidth]{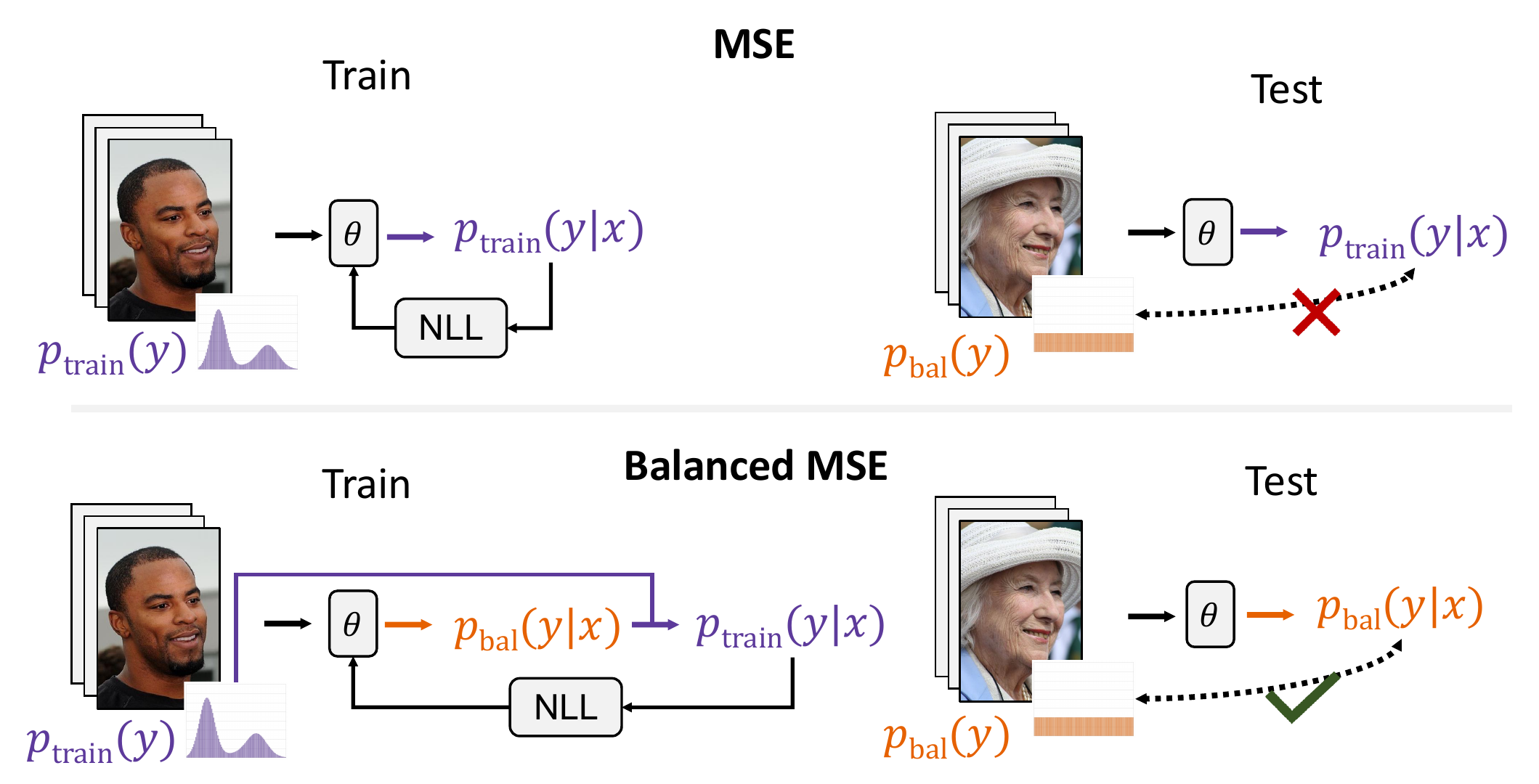}
	\caption{Comparison between MSE and Balanced MSE. MSE is equivalent to NLL on a prediction distribution, where the regressor $\boldsymbol{\theta}$'s prediction is the distribution mean. MSE lets the regressor model $\pyxtrain$, which is not suitable to infer on the test set due to a shift between the training label distribution $\pytrain$ and the balanced test label distribution $\pybal$. In comparison, Balanced MSE leverages $\pytrain$ to make a statistical conversion from $\pyxbal$ to $\pyxtrain$, thus allowing the regressor to model the desired $\pyxbal$ by still minimizing NLL of $\pyxtrain$.} \label{fig:illustration}
	\vspace{-15pt}
\end{figure}

\begin{figure*}
\centering
    \vspace{-10pt}
	\includegraphics[width=1.\textwidth]{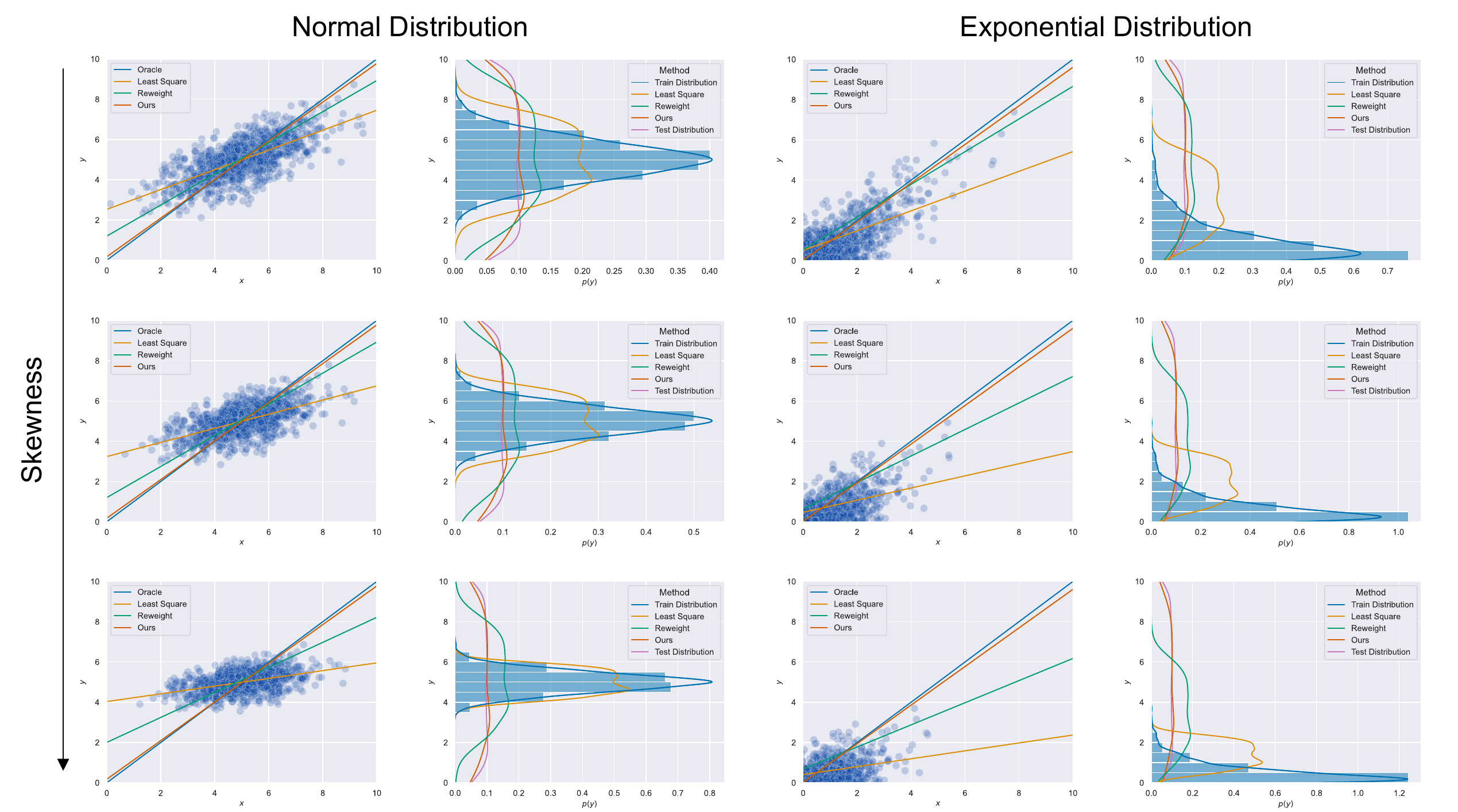}
	\vspace{-16pt}
	\caption{Comparison of \name{} and existing methods on a 1-D imbalanced linear regression synthetic benchmark. Column 1 and 3 are visualizations of regression results: points represent training data, $x$ is input and $y$ is label; $(x, y)$ is noisily generated by an oracle linear relation (in blue) and is artificially label-imbalanced; other lines represent different regressors, the closer to the oracle the better. Column 2 and 4 are visualizations of label distributions: blue shaded histogram represents the training label distribution $\pytrainuni$, it gets more skewed from top to bottom; purple histogram represents the test label distribution, which is balanced; other histograms are the marginal label distributions predicted by different regressors on the test set, the closer to the test distribution the better. Although reweighting (in green) is closer to the oracle (in blue) compared with least square (in yellow), it suffers a larger error when $\pytrainuni$ gets more skewed. Our method (in red), \name{}, makes the estimation closest to the oracle and has a uniform marginal label distribution on the test set.} \label{fig:case_study_1d_skewness}
	\vspace{-15pt}
\end{figure*}

Visual regression, where models learn to predict continuous labels, is one of the most fundamental tasks in machine learning. However, in real-world applications, data imbalance is widely encountered, hurting the model's generalizability and fairness. For example, age estimation predicts people's age from their visual appearance, where age is a continuous label. In practice, most of the training images are from adults, while very few images are from children and senior adults. As a result, models trained from such an imbalanced dataset can have inferior performance on under-represented groups~\cite{yang2021delving}. Therefore, imbalanced regression gains increasing research attention. A fresh imbalanced regression benchmark~\cite{yang2021delving} in the modern deep learning context has been curated recently as well. Compared with imbalanced and long-tailed classification~\cite{buda2018systematic, Liu2019LargeScaleLR, Gupta2019LVISAD} that studies categorical labels, imbalanced regression focuses on continuous labels, which can be boundless, high-dimensional, and hence more challenging. 

Unlike imbalanced classification that has been widely discussed~\cite{kang2019decoupling, zhou2020bbn, wang2020long, hong2021disentangling}, imbalanced regression is under-explored. Previous works~\cite{branco2017smogn, torgo2013smote} focus on synthesizing samples for rare labels, which have limited feasibility in modern deep learning where inputs are always high-dimensional. Recent research~\cite{yang2021delving, steininger2021density} focuses on loss reweighting. Reweighting assigns larger loss weights to rare samples and smaller loss weights to frequent samples. ~\cite{yang2021delving, steininger2021density} estimates the training label distributions using kernel density estimation (KDE) and reweight losses accordingly. However, prior works~\cite{byrd2019effect, xu2021understanding} show that reweighting has limited effectiveness on imbalanced classification. In a following case study, we validate this finding in imbalanced regression as well. To sum up, \textit{imbalanced regression is still in an early stage and lacks an effective approach}.

To mitigate the gap, we present a statistically principled loss function, \textbf{\name{}}, for imbalanced regression. We revisit Mean Square Error (MSE), the standard loss function in regression, from a statistical view. We identify that MSE carries the label imbalance into predictions, which leads to inferior performance on rare labels. We propose Balanced MSE to restore a balanced prediction by leveraging the training label distribution prior to make a statistical conversion. Moreover, we provide various implementation options for Balanced MSE, including the one that estimates the training label distribution online and requires no additional prior knowledge, making Balanced MSE applicable to different real-world scenarios.

\name{} shows clear advantages over existing methods both theoretically and practically. As a motivating example, we compare \name{} with reweighting using an 1-D linear regression synthetic benchmark shown in \autoref{fig:case_study_1d_skewness}. Regressors trained with \name{} show a consistent performance that is invariant to the skewness of the training label distribution. On the contrary, reweighting suffers from a significantly larger prediction error when the training label distribution gets more skewed.

We further demonstrate \name{}'s empirical success on existing real-world benchmarks~\cite{yang2021delving}, including age estimation and depth estimation. Note that existing imbalanced regression benchmarks only consider uni-dimensional label space, \eg, age and depth. However, labels sometimes have more than one dimension in real-world applications. To close the gap, we propose a new multi-dimensional imbalanced regression benchmark on Human Mesh Recovery (HMR)~\cite{hmrKanazawa17}, which is an important task that estimates 3D human meshes from monocular images. We extend the standard metrics of HMR (e.g., mean per joint position error (MPJPE)) to balanced metrics so that we evaluate the regression performance on human meshes with different rarity fairly. We call the new imbalanced regression benchmark Imbalanced HMR (\benchmark{}). We show that \name{} delivers strong empirical results on both uni- and multi-dimensional benchmarks. \textit{To the best of our knowledge, Balanced MSE is the first general solution to high-dimensional imbalanced regression in modern context.}

In summary, our contributions are three-fold:
\textbf{1)} We identify the ineffectiveness of MSE in imbalanced regression and propose a statistically principled loss function, \name{}, that leverages the training label distribution prior to restore a balanced prediction.
\textbf{2)}  We devise various implementation options of Balanced MSE to tackle different real-world scenarios, including the one that estimates the training label distribution online and requires no prior knowledge beforehand. 
\textbf{3)} We propose a new multi-dimensional benchmark \benchmark{}, and show that \name{} achieves state-of-the-art performance on both uni- and multi-dimensional real-world benchmarks.

%% file: sections/2_related_works.tex
\section{Related Works}
\noindent\threesubsection{Imbalanced \& Long-Tailed Classification}
Many techniques have been explored for imbalanced \& long-tailed classification, for example, resampling~\cite{chawla2002smote, he2009learning, kim2020m2m, chu2020feature} and reweighting~\cite{huang2016learning, cui2019class, jamal2020rethinking, cao2019learning}. Here, we focus on the logit adjustment techniques, which are the most relevant to this work.
Recent works~\cite{ren2020balanced, tian2020posterior, menon2020long} show that modifying the logits in the mapping function, e.g., Softmax or Sigmoid, by an offset proportional to $\log\pytrainuni$ gives the Bayes-optimal estimation of the $\pyxbaluni$. The logit adjustment techniques can work as either a train-time loss function or a test-time adjustment. \cite{wang2021seesaw} further develops an online version that accumulates the statistics of label distribution during training instead of requiring statistics of all training labels ahead of time.

\noindent\threesubsection{Imbalanced Regression}
Imbalanced regression is relatively under-explored. Earlier works ~\cite{torgo2013smote, branco2017smogn} focus on resampling and synthesizing new samples for rare labels.
Further work~\cite{branco2018rebagg} ensembles regressors trained under different resampling policies. Extending their method towards high-dimensional observations like images is non-trivial. Recent research \cite{yang2021delving, steininger2021density} proposes to estimate the empirical training distribution with KDE and then apply the standard reweighting technique.  \cite{yang2021delving} proposes a feature level smoothing as well, which is complementary to this work. 

%% file: sections/3_method.tex
\section{Methodology}
\subsection{Problem Setting}\label{sect:Prelimearies}

We study a regression task. We consider input $\boldsymbol{x} \in X$ and label $\boldsymbol{y} \in Y=\rspace^{d}$. Different from ~\cite{yang2021delving} which only discusses uni-dimensional ($d=1$) regressions, we discuss multi-dimensional ($d>1$) regressions in this paper as well. 

Normally, both the training set and test set are drawn from the same joint distribution. However, when the label distribution is highly skewed, a model may learn a trivial solution by always predicting the frequent labels. The trivial model will still have a low error rate on the test set~\cite{menon2020long}. To address the issue, either a balanced evaluation metric or a balanced test set is employed to fairly evaluate a model's performance on samples with different rarities. Moreover, it can be shown that using a balanced metric on an arbitrary test set is equivalent to using an overall metric on a balanced test set that hypothetically exists~\cite{brodersen2010balanced}. 

Therefore, imbalanced regression assumes that the training set and test set are drawn from different joint distributions, $p_{\textup{train}}(\boldsymbol{x}, \boldsymbol{y})$ and $p_{\textup{bal}}(\boldsymbol{x}, \boldsymbol{y})$ respectively, where the training set's label distribution $\pytrain$ is skewed and the balanced test set's label distribution $\pybal$ is uniform~\cite{yang2021delving}. The label-conditional probability $\pxy$ is assumed to be the same in both training and testing. Instead of learning $\pyxtrain$, imbalanced regression's goal is to estimate $\pyxbal$ to better perform on the balanced test set. A similar setting is generally adopted in imbalanced 
classification literature~\cite{cao2019learning, menon2020long, ren2020balanced, hong2021disentangling} as well.

\subsection{Revisiting Mean Square Error}
In this section, we revisit Mean Square Error (MSE) from a statistical view. MSE loss is the most commonly used loss function in regression. For a predicted label $\ypred$ and a target label $\boldsymbol{y}$, the MSE loss can be written as 
\begin{align}\label{eq:mse}
\textup{MSE}(\boldsymbol{y}, \ypred) = \norm{\boldsymbol{y} - \ypred}^2_2,
\end{align}
where $\norm{.}_2$ denotes the L2 norm. It is well known that minimizing MSE can be equivalent to maximum likelihood estimation in regression~\cite{nix1994estimating}. The prediction of a regressor $\ypred$ can be considered as the mean of a noisy prediction distribution, which is modeled as a Gaussian distribution in the classic probabilistic interpretation~\cite{mccullagh1989generalized}:
\begin{align}\label{eq:Gaussian}
\pyxtheta = \gau(\boldsymbol{y};\ypred, \Sigmapred),
\end{align}
where $\boldsymbol{\theta}$ is the regressor's parameter, $\ypred$ is the regressor's prediction and $\sigmapred$ is the scale of an i.i.d. error term $\epsilon \sim \gau(0, \Sigmapred)$. It is easy to show that MSE equals to the Negative Log Likelihood (NLL) loss of the prediction distribution $\pyxtheta$~\cite{nix1994estimating}. Therefore, a regressor trained using MSE in fact learns to model $\pyxtrain$.

However, as mentioned in the problem setting, we are interested in estimating $\pyxbal$ instead of $\pyxtrain$. Due to a shift from the long-tailed training distribution $\pytrain$ to the balanced test distribution $\pybal$, there is a mismatch between $\pyxtrain$ and $\pyxbal$.
By Bayes' Rule, we have $\pyxtrain \propto \pxy\cdot\pytrain$ and $\pyxbal \propto \pxy\cdot\pybal$.
By change of variables, we have:
\begin{align}\label{eq:change_variable}
    \frac{\pyxtrain}{\pyxbal} & \propto \frac{\pytrain}{\pybal} 
\end{align}
\autoref{eq:change_variable} quantifies that the ratio between $\pyxtrain$ and $\pyxbal$ is proportional to $\pytrain$, which is lower when a label rarely appears in the training set. Therefore, \textit{a regressor trained with MSE will underestimate on rare labels.}

Although this mismatch is a well-known observation in imbalanced classification~\cite{tian2020posterior, menon2020long, ren2020balanced, hong2021disentangling}, we are the first to address the mismatch in imbalanced regression. Different from classification tasks that model prediction distribution explicitly by Softmax scores, regression tasks model the prediction distribution implicitly: only the mean of the prediction distribution plays a role in both training and testing. Therefore, the probabilistic meaning of imbalanced regression has been constantly overlooked by existing research. Our work is an initial attempt to rethink imbalanced regression in a statistical framework. We will show how the statistical insight sheds light on imbalanced regression in the following sections.

\subsection{\name{}}\label{sect:Bayesian Posterior Debiasing}

We propose Balanced MSE to restore $\pyxbal$. First, we discuss a statistical conversion from $\pyxbal$ to $\pyxtrain$ using the training label distribution $\pytrain$.
\begin{theorem}[Statistical Conversion]\label{theorem:PD}
Let $p_{\textup{train}}(\boldsymbol{x}, \boldsymbol{y})$ be the training distribution where $\pytrain$ is imbalanced and $p_{\textup{bal}}(\boldsymbol{x}, \boldsymbol{y})$ be the balanced test distribution where $\pybal$ is uniform. $p_{\textup{train}}(\boldsymbol{x}, \boldsymbol{y})$ and $p_{\textup{bal}}(\boldsymbol{x}, \boldsymbol{y})$ have the same label-conditional distribution $\pxy$. $\pyxtrain$ can always be expresed by $\pyxbal$ and $\pytrain$ as:
\begin{align}\label{eq:inst_reg}
    \pyxtrain = \frac{\pyxbal \cdot \pytrain}{\int_Y\pyprimexbal\cdot\pyprimetrain\dyprime}.
\end{align}

\end{theorem}
The proof can be found in the supplementary materials. \autoref{theorem:PD} allows us to estimate $\pyxbal$ by minimizing the NLL loss of $\pyxtrain$. Specifically, we let the regressor directly estimate the desired $\pyxbal$, \ie, 
\begin{align}
    \pyxbaltheta=\gau(\boldsymbol{y};\ypred, \Sigmapred)
\end{align}
As illustrated in \autoref{fig:illustration}, in training, we first predict $\pyxbaltheta$, convert it into  $\pyxtraintheta$ using \autoref{eq:inst_reg} and then compute the NLL loss to update $\boldsymbol{\theta}$; in testing, we skip the conversion and directly output the regressor's prediction $\pyxbaltheta$. We name the NLL loss of the converted conditional probability as Balanced MSE. 
\begin{definition}[Balanced MSE]
For a regressor's prediction $\ypred$, and a training label distribution prior $\pytrain$, the Balanced MSE loss is defined as:
\begin{equation}
\begin{split}\label{eq:balanced_mse}
L &= -\log \pyxtraintheta\\
&=-\log\frac{\pyxbaltheta \cdot \pytrain}{\int_Y\pyprimexbaltheta\cdot\pyprimetrain\dyprime}\\
&\cong-\log \gau(\boldsymbol{y};\ypred,\Sigmapred) \\
&+ \log\int_Y\gau(\yprime;\ypred,\Sigmapred)\cdot\pyprimetrain\dyprime,
\end{split}
\end{equation}
where $\cong$ hides a constant term $-\log\pytrain$.
\end{definition}
\name{} has two parts: the first part is equivalent to the standard MSE loss, and the second part is a new balancing term, where an integral needs to be computed.
We show in the supplementary materials that the new balancing term equals a constant when the training label distribution $\pytrain$ is uniform. Therefore, the standard MSE loss can be viewed as a special case of \name{}. 

\name{} closes the distribution mismatch between training and testing, thus being a statistically principled loss function for imbalanced regression. In the following sections, we discuss 1) \name{}'s connection with imbalanced classification and 2) how to implement \name{} in practice.

\subsection{Connection with Imbalanced Classification}\label{sect:Instantiation on the imbalanced Classification}
We show that Balanced MSE has an underlying connection with existing solutions in imbalanced classification. \autoref{theorem:PD} is true not only in imbalanced regression but also in imbalanced classification. In imbalanced classification, the label space $Y$ is one-dimensional and discrete, the integral on $Y$ can be written into summation, \autoref{eq:inst_reg} becomes:
\begin{align}\label{eq:inst_cls}
    \pyxtrainuni = \frac{\pyxbaluni \cdot \pytrainuni}{\sum_{y^\prime\in Y}\pyprimexbaluni \cdot\pyprimetrainuni}.
\end{align}
Usually, Softmax is employed to convert model outputs into a prediction distribution in classification. When using Softmax to express the desired $\pyxbaluni$, we have:
\begin{align}\label{eq:softmax}
\pyxbalthetauni = \frac{ \exp({\eta[y]})}{ \sum_{y^\prime \in Y} \exp({\eta[y^\prime]})},
\end{align}
where $\eta[y]$ is the model's output on class $y$.
Plugging the Softmax expression in \autoref{eq:softmax} into \autoref{eq:inst_cls}, we have
\begin{align}\label{eq:logit_adjustment}
\pyxtrainthetauni = \frac{ \exp({\eta[y]}) \cdot \pytrainuni}{ \sum_{y^\prime\in Y} \exp({\eta[y^\prime]}) \cdot \pyprimetrainuni},
\end{align}
which achives the same form as the logit adjustment techniques in imbalanced classification literature~\cite{ren2020balanced, menon2020long, hong2021disentangling}.
Therefore, \name{} and the logit adjustment techniques can be viewed as two different instantiations of \autoref{theorem:PD} on imbalanced regression and imbalanced classification respectively. \textit{Our work provides a unified statistical view of both imbalanced classification and regression for the first time}. We hope that the unified perspective can help future research to further bridge the two tasks.

\subsection{Implementation Options}
We discuss how to implement Balanced MSE in practice. The integral in \name{} (\autoref{eq:balanced_mse}) can be difficult to compute. In the following sections, we provide both closed-form options and numerical options to calculate the integral. Particularly, the option Batch-based Monte-Carlo (BMC) requires no prior knowledge of the training label distribution and hence can be more generally deployed in real-world applications.

\subsubsection{Closed-form Options}
In this section, we aim to find a closed-form expression for the integral $\int_Y\gau(\boldsymbol{y};\ypred,\Sigmapred)\cdot\pytrain\dy$. The main challenge is how to express $\pytrain$ to make the integral tractable. Here, we discuss a viable option, which is to express $\pytrain$ as a Gaussian Mixture Model (GMM).

\noindent\threesubsection{GMM-based Analytical Integration (GAI)}
The advantage of employing GMM is the fact that the product of two Gaussians is an unnormalized Gaussian. Concretely, let us have $\pytrain$ expressed by a Gaussian Mixture:
\begin{align}\label{eq:gmm}
    \pytrain=\sum_{i=1}^K\phi_i\gau(\boldsymbol{y};\boldsymbol{\mu}_i, \boldsymbol{\Sigma}_i),
\end{align}
where $K$ is the number of Gaussian components, $\phi, \boldsymbol{\mu}, \boldsymbol{\Sigma}$ are the weights, means and covariances of the GMM.
Since the product of two Gaussians is an unnormalized Gaussian, we have:
\begin{equation}
\begin{split}\label{eq:gmm_prod}
    \int_Y&\gau(\boldsymbol{y};\ypred, \Sigmapred)\cdot \sum_{i=1}^K\phi_i\gau(\boldsymbol{y};\boldsymbol{\mu}_i, \boldsymbol{\Sigma}_i)\dy  \\ &=\sum_{i=1}^K{\phi}_iS_i\int_Y\gau(\boldsymbol{y};\tilde{\boldsymbol{\mu}}_i, \tilde{\boldsymbol{\Sigma}}_i)\dy.
\end{split}
\end{equation}
where $S, \tilde{\boldsymbol{\mu}}, \tilde{\boldsymbol{\Sigma}}$ are the norms, means, and covariances of the new unnormalized Gaussian. Now, the integral is on Gaussian distribution and can be trivially solved. We leave the detailed derivation in the supplementary material. The final loss form is:
\begin{equation}
\begin{split}\label{eq:gmm_loss_multi}
L &= -\log \gau(\boldsymbol{y};\ypred,\Sigmapred) \\
&+ \log\sum_{i=1}^K\phi_i\cdot\gau(\ypred;\boldsymbol{\mu}_i, \boldsymbol{\Sigma}_i + \Sigmapred).
\end{split}
\end{equation}

\subsubsection{Numerical Options}
The closed-form solution above imposes a constraint on the modeling of $\pytrain$. However, in modern deep learning tasks, $\pytrain$ could be very high-dimensional and has a complex underlying distribution. With the constraint on the distribution modeling, analytically expressing $\pytrain$ could be challenging. Therefore, we discuss a few numerical approaches, which could be more generally applicable to all types of label data but could bear a larger variance in optimization. In essence, we use Monte Carlo Method (MCM) to approximate $\pytrain$:
\begin{equation}
\begin{split}\label{eq:mcm}
&\int_Y\gau(\boldsymbol{y};\ypred,\Sigmapred)\cdot\pytrain\dy \\
&= \ev_{\boldsymbol{y}\sim\pytrain}[\gau(\boldsymbol{y};\ypred,\Sigmapred)] \\
&\approx \frac{1}{N}\sum_{i=1}^N \gau(\boldsymbol{y}_{(i)};\ypred,\Sigmapred).
\end{split}
\end{equation}

\noindent\threesubsection{Batch-based Monte-Carlo (BMC)}
BMC requires no prior knowledge on $\pytrain$. It treats all labels in a training batch as random samples from $\pytrain$. For labels in a training batch $B_{\boldsymbol{y}} = \{\boldsymbol{y}_{(1)}, \boldsymbol{y}_{(2)}, ... \boldsymbol{y}_{(N)}\}$, the loss will be:
\begin{equation}
\begin{split}\label{eq:mcm_batch}
L &=  -\log \gau(\boldsymbol{y};\ypred,\Sigmapred) \\
 &+\log\sum_{i=1}^N\gau(\boldsymbol{y}_{(i)};\ypred, \Sigmapred).
\end{split}
\end{equation}
Furthermore, BMC in \autoref{eq:mcm_batch} can be rewritten like Softmax with temperature:
\begin{align}\label{eq:bmc_loss}
L = -\log \frac{\exp(-\norm{\ypred - \boldsymbol{y}}^2_2/ \tau)}{ \sum_{\yprime \in B_{\boldsymbol{y}}} \exp(-\norm{\ypred-\yprime}^2_2 / \tau)},
\end{align}
where $\tau=2\sigmapred^2$ is a temperature coefficient. 

BMC is easy to implement. Interestingly, its form in \autoref{eq:bmc_loss} is equivalent to \textit{classifying within a batch}, and shows similarity to contrastive loss functions used in self-training~\cite{he2020momentum, chen2020simple}. The similarity could potentially be connected with self-training's effectiveness in imbalanced learning~\cite{yang2020rethinking, kang2021explore}, which we leave for future discussion.

\noindent\threesubsection{Bin-based Numerical Integration (BNI)}
Although the "bin" based idea mainly applies to uni-dimensional label space, it allows us to leverage recent progress on estimating label densities using KDE~\cite{yang2021delving,steininger2021density}. These prior works first divide the label space into evenly distributed bins, then use KDE to estimate the $\pytrain$ at the bin centers. We may directly use their results to make a numerical integration. For $N$ bin centers $\{\boldsymbol{y}_{(1)}, \boldsymbol{y}_{(2)}, ..., \boldsymbol{y}_{(N)}\}$, the loss is:
\begin{equation}
\begin{split}\label{eq:mcm_batch_loss}
L &= -\log \gau(\boldsymbol{y};\ypred,\Sigmapred) \\
 &+ \log\sum_{i=1}^N\pysampletrainwoeq\cdot\gau(\boldsymbol{y}_{(i)};\ypred, \Sigmapred).
\end{split}
\end{equation}

\subsubsection{Finding Optimal Noise Scale}
Unlike the standard MSE loss, the noise scale $\sigmapred$ makes a difference in the proposed method. Locating an optimal noise scale is thus important. A hyper-parameter search on $\sigmapred$ will be affordable given that $\sigmapred$ is defined in $\rspace_+$ and bounded by the square root of train-time and test-time MSEs. However, in this paper, instead of using hyper-parameter search, we jointly optimize $\sigmapred$ with $\ypred$ during model training.

We observe that we can obtain near-optimal $\sigmapred$ by simply setting $\sigmapred$ as a learnable parameter. A comparison between using the ground truth noise scale and using the jointly learned $\sigmapred$ is shown in the supplementary material. Therefore, no additional hyper-parameter tuning is required by Balanced MSE, making Balanced MSE more friendly to practitioners. We adopt the joint optimization paradigm in all empirical analyses unless specified.

%% file: sections/4_experiments.tex
\begin{figure*}[!ht]
\centering
	\includegraphics[width=\textwidth]{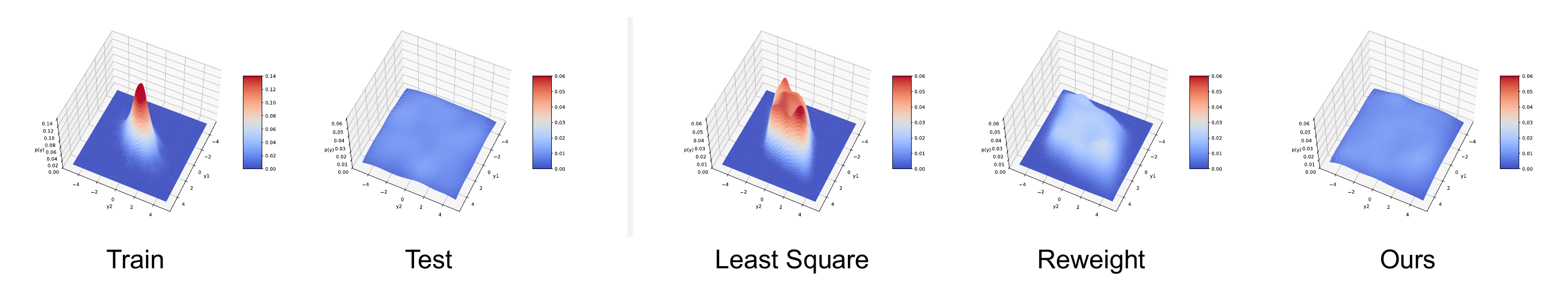}
	\vspace{-20pt}
	\caption{Comparison of marginal label distributions on 2D linear regression. Least square and reweighting show visible bias towards the high-frequency area around the center. In comparison, \name{} achieves the closest marginal label distribution to the uniform test distribution.
	} \label{fig:case_study_2d}
	\vspace{-10pt}
\end{figure*}

\begin{figure*}[!ht]
\centering
	\includegraphics[width=\textwidth]{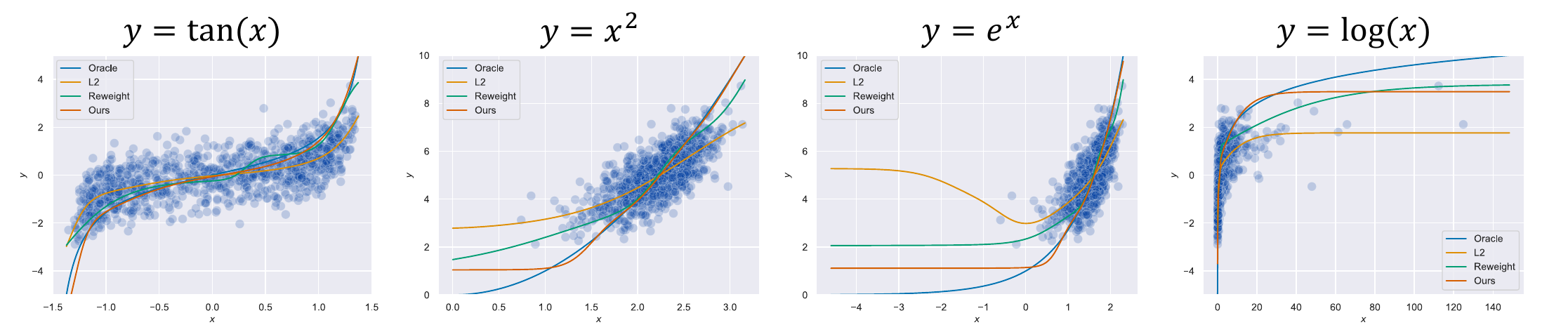}
	\vspace{-25pt}
	\caption{Qualitative comparison for nonlinear regression. Four nonlinear functions are studied. \name{} (in red) gives the closest estimation to the oracle (in blue). }
	\label{fig:case_study_1d_nonlinear}
	\vspace{-10pt}
\end{figure*}

\section{Experiments}

\subsection{Synthetic Benchmarks}

\input{tables/4_comparison_imdb}

We construct a simple one-dimensional linear imbalanced regression dataset, with the training label distribution being normal or exponential and skewed to various extents. We train a one-layer linear regressor on the imbalanced training set and test on a uniform test set with no additive noise. We compare three types of regressors: a least-square estimator, a linear regressor inversely reweighted by the true $\pytrain$ as described in \cite{yang2021delving}, and \name{}'s closed-form option GAI with true noise scale. We show the visualized results in \autoref{fig:case_study_1d_skewness}.
We observe that the reweighted regressor shows increasingly larger error when the training distribution becomes more skewed. In comparison, \name{} gives an accurate estimation that is robust to different levels of skewness.

We further compare the three methods on a two-dimensional regression. The training label distribution is set as a Multivariate Normal (MVN) distribution. We visualize the marginal label distributions in~\autoref{fig:case_study_2d}, where \name{} achieves a marginal label distribution closest to uniform. For nonlinear regressions, \name{} achieves a consistent effectiveness as well, as shown in \autoref{fig:case_study_1d_nonlinear}. We provide another experiment on random seeds in the supplementary material to demonstrate \name{}'s robustness to noise.

Despite recent works~\cite{yang2021delving, steininger2021density} focusing on estimating the training label distribution, our synthetic benchmark shows that \textit{the bottleneck of existing techniques is reweighting}. Even given the true label distribution, reweighting fails to find the optimal estimator in all settings. Our conclusion aligns with recent research that shows reweighting's incapability on imbalanced classification~\cite{byrd2019effect, xu2021understanding}. In comparison, our proposed \name{} is robust to different skewness of the training distribution and noise, meanwhile applicable to nonlinear and multi-dimensional regressions.

We provide quantitative results for the above described synthetic benchmark in the supplementary material, where we compare different implementation options and choices of noise scale as well.  The results show that the numerical option achieves comparable results with closed-form options. Moreover, the jointly-optimized noise scale achieves near-optimal results in most cases.

\input{tables/4_comparison_nyu}

\begin{figure*}
\centering
	\includegraphics[width=1.\textwidth]{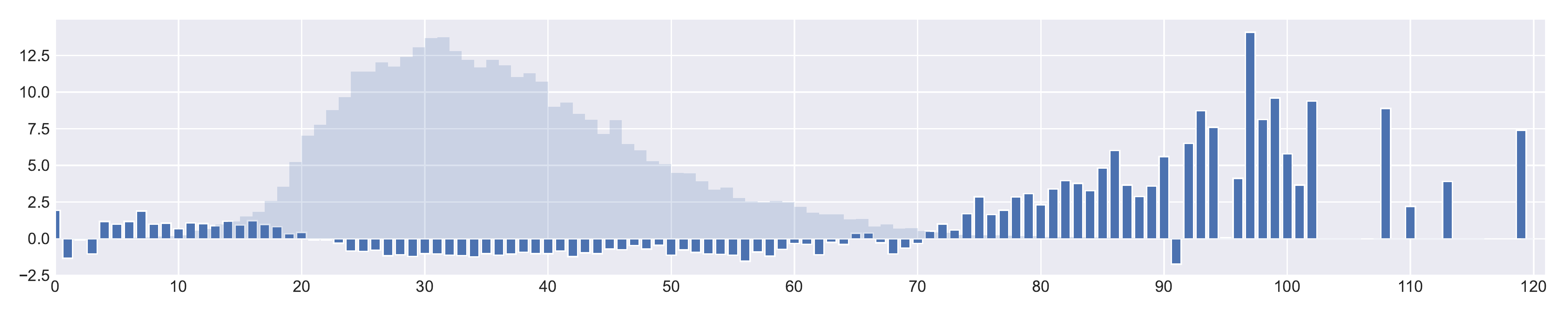}
    \vspace{-20pt}
	\caption{\name{}'s bMAE gain over the baseline. The light blue area in the background shows the training label histogram of IMDB-WIKI-DIR. \name{} improves the performance on tail labels  (age $<$ 20 and $>$ 70) substantially.} \label{fig:bmae_diff}
\end{figure*}

\begin{figure*}[t]
\centering
	\includegraphics[width=1.\textwidth]{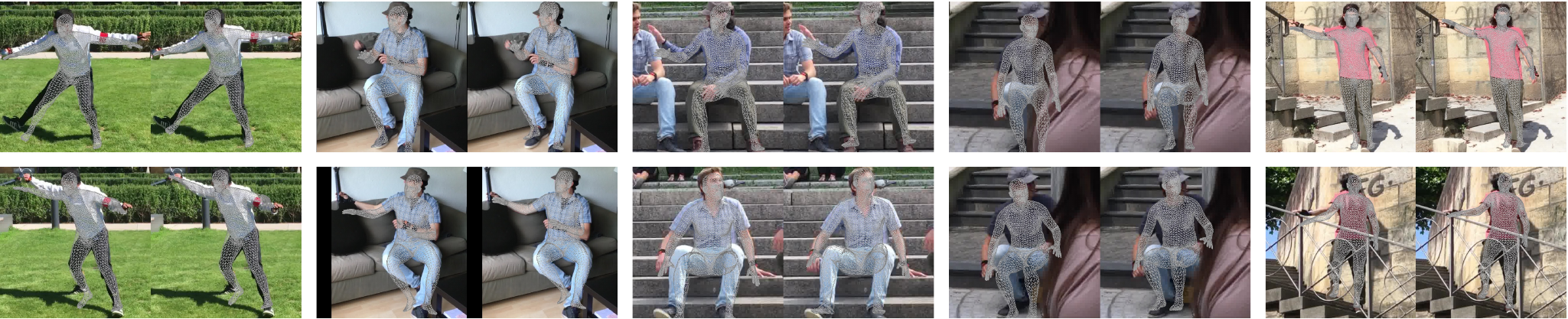}
	\vspace{-20pt}
	\caption{Qualitative comparison of \name{} and the baseline, SPIN-RT. Left: SPIN-RT. Right: \name{}. We observe that the baseline's predictions are less stretched out. They bias towards the mean pose, particularly for poses like raising arms and bending legs. In comparison, our method effectively eliminates the bias and recovers rare poses.} \label{fig:smpl_vis}
	\vspace{-2pt}
\end{figure*}

\subsection{Real-World Benchmarks}

\subsubsection{Datasets and Settings}
\threesubsection{Age and Depth Estimation}
We select two representative tasks from~\cite{yang2021delving}'s DIR benchmark. We estimate ages from face images on the IMDB-WIKI-DIR dataset and estimate depth maps from images of indoor scenes on the NYUD2-DIR dataset.

\noindent\threesubsection{Imbalanced Human Mesh Recovery (IHMR)}
IHMR is a new, multi-dimensional imbalanced regression benchmark. We estimate human meshes from images, where the mesh is represented by a parametric human model known as SMPL~\cite{loper2015smpl}. Typically, SMPL model has two parameters: $\boldsymbol{\theta}\in\rspace^{24\times3}$ represents the rotation of 23 body joints and 1 global orientation and $\boldsymbol{\beta} \in \rspace^{10}$ represents the 10 PCA components for body shape. Therefore, the label space of IHMR is multi-dimensional. Aligned with recent works \cite{rong2020chasing}, we observe that the distribution of human meshes is long-tailed. We show a visualization of training distribution in the supplementary material. Following \cite{kolotouros2019learning}, we train on a combination of 3D and 2D human datasets and test on an in-the-wild 3D dataset. Detailed settings can be found in the supplementary material.

\subsubsection{Evaluation Metrics}
The DIR benchmark~\cite{yang2021delving} uses primarily overall metric, \eg, Mean Absolute Error (MAE) to report the performance on the benchmark. This is on the assumption that the test dataset is perfectly balanced. However, we observe visible tails in IMDB-WIKI-DIR's test set as shown in the supplementary material. To fairly measure the model's performance on tail labels, we follow the idea of balanced metrics~\cite{brodersen2010balanced}, and divide the label space into a finite number of even sub-regions, compute the average inside the sub-regions, and take the mean overall sub-regions. We name it "balanced-" ("b-") metric, \eg, bMAE. 

\noindent\threesubsection{Age \& Depth Estimation}We primarily report bMAE on IMDB-WIKI-DIR. NYUD2-DIR's test set is balanced, we follow \cite{yang2021delving} and report RMSE.

\noindent\threesubsection{Imbalanced Human Mesh Recovery (IHMR)} We extend HMR's evaluation metrics to balanced metrics in IHMR. We evenly divide the label space into 100 sub-regions according to their vertex-based distances to the mean parameter and compute balanced metrics as described above. Following~\cite{rong2020chasing}, we primarily report balanced mean per-vertex position error (bMPVPE). We also report balanced mean per-joint position error (bMPJPE) and balanced Procrustes-aligned mean per joint position error (bPA-MPJPE). 
We include the "tail 5\%" metric and the "tail 10\%" metric to show performance on extreme poses as well.

\input{tables/4_effectiveness_on_hmr}

\subsubsection{Comparison Results}

\autoref{tab:comparison_w_sota_imdb} shows a comparison with state-of-the-art (SOTA) methods on age estimation. Regressor Re-Training (RRT)~\cite{yang2021delving} first trains the feature extractor normally and retrain the last linear layer using inverse re-weighting. RRT+LDS is an improved version of RRT, where training label distribution is estimated using Label Distribution Smoothing~\cite{yang2021delving}. RRT and RRT+LDS are the best performing regressor learning methods on IMDB-WIKI-DIR in~\cite{yang2021delving}'s benchmark. Balanced MSE substantially outperforms the previous methods. Notably, the BMC option outperforms SOTAs with a large margin without relying on the pre-processed training label distribution. We further analyze the bMAE gain in \autoref{fig:bmae_diff} and observe an effective trade-off between frequent and rare labels towards a balanced estimation. 
Note that we do not include Feature Distribution Smoothing (FDS) in the comparison since it works on feature learning and should be complementary to our method.

\autoref{tab:comparison_w_sota_nyu} shows comparison with SOTA on depth estimation. Note that depth map has an inter-pixel dependency, the pixel-wise error $\sigmapred$ can be under-estimated and the BMC can give an inaccurate estimation to $\pytrain$. We set a fixed $\sigmapred$ to 1, and use BNI for numerical option evaluation. Compared with the SOTA, both closed-form and numerical implementations achieve clear improvements. 

\autoref{tab:comparison_w_sota_smpls} shows a comparison between \name{} and existing HMR methods. \name{} outperforms the baseline by a large margin on the main metric bMPVPE (-3.4). We show the qualitative comparison in \autoref{fig:smpl_vis}. PM-Net~\cite{rong2020chasing} achieves better results on tail-5\% bMPVPE, by designing prototypes and adaptively selecting them as the initialization for SMPL regression. It is noteworthy that PM-Net improves the regression initialization and should be complementary to our method.

%% file: tables/4_comparison_imdb.tex
\begin{table*}
\small
 \centering
 \caption{Comparison experiment on IMDB-WIKI-DIR. $\dagger$: MAE metric reported in \cite{yang2021delving}. Best results are bolded.}
 \vspace{-10pt}
\label{tab:comparison_w_sota_imdb}
\begin{tabular}{lrrrrrrrr}

    \toprule
     & 
    \multicolumn{4}{c}{bMAE$\downarrow$}
    &
    \multicolumn{4}{c}{MAE$\downarrow$} \\ 
    
    \cmidrule(lr){2-5} 
    \cmidrule(lr){6-9} 

Method & All & Many & Med. & Few  & All & Many & Med. & Few \\
\midrule
Vanilla$^\dagger$  & 13.92 & 7.32 & 15.93 & 32.78 & 8.06 & 7.23  & 15.12 & 26.33 \\
\midrule
RRT$^\dagger$  &  13.12 & \textbf{7.27}  & 14.03  & 30.48 & 7.81 & \textbf{7.07} & 14.06 & 25.13\\
RRT+LDS$^\dagger$  & 13.09 & 7.30 & 14.05 & 30.26 & \textbf{7.79} & 7.08 &  13.76 & 24.64 \\
\midrule
Ours (BMC) & 12.69 & 7.59 & 12.90 & 28.28  & 8.08 &7.52  & 12.47 & 23.29 \\ 
Ours (GAI) & \textbf{12.66} & 7.65 & \textbf{12.68} & \textbf{28.14} & 8.12 &  7.58 & \textbf{12.27} &\textbf{23.05}\\

\bottomrule
\end{tabular}
\end{table*}

%% file: tables/4_comparison_nyu.tex
\begin{table*}
\small
 \caption{Comparison experiment on NYUD2-DIR. $\dagger$: reported in \cite{yang2021delving}. 
  Best results are bolded.
 }
 \vspace{-10pt}
\label{tab:comparison_w_sota_nyu}
 \centering
\begin{tabular}{lrrrrrrrr}

    \toprule
    &
    \multicolumn{4}{c}{RMSE$\downarrow$} 
    &
    \multicolumn{4}{c}{$\delta_1\uparrow$} \\ 
    \cmidrule(lr){2-5} 
    \cmidrule(lr){6-9} 

Method & All & Many & Med. & Few & All & Many & Med. & Few\\
\midrule
Vanilla$^\dagger$ & 1.477 & \textbf{0.591} & 0.952 & 2.123 & 0.677 & \textbf{0.777} & 0.693 & 0.570 \\
Vanilla + LDS$^\dagger$  &  1.387 & 0.671 & 0.913 & 1.954 & 0.672 & 0.701 & 0.706 & 0.630 \\
\midrule
Ours (BNI) & 1.283  & 0.787  & \textbf{0.870} & 1.736 & 0.694 & 0.622 & \textbf{0.806} & \textbf{0.723} \\
Ours (GAI) & \textbf{1.251}  & 0.692  & 0.959 & \textbf{1.703} & \textbf{0.702} & 0.676 & 0.734 & 0.715 \\
\bottomrule
\end{tabular}
\end{table*}

%% file: tables/4_effectiveness_on_hmr.tex
\begin{table*}[t!]
\centering
\small
\caption{Comparison experiment on Imbalanced Human Mesh Recovery. $\dagger$: reported in \cite{rong2020chasing}. SPIN-RT: keep the SPIN's feature extractor fixed and retrain the last linear regression layers. Best results are bolded.
\vspace{-10pt}
} \label{tab:comparison_w_sota_smpls}
\begin{tabular}{lrrrrrrrrr}

    \toprule
     & 
    \multicolumn{3}{c}{bMPVPE$\downarrow$} 
    &
    \multicolumn{3}{c}{bMPJPE$\downarrow$} 
    &
    \multicolumn{3}{c}{bPA-MPJPE$\downarrow$} \\ 
    
    \cmidrule(lr){2-4} 
    \cmidrule(lr){5-7} 
    \cmidrule(lr){8-10} 
    
Method & All & 10\% & 5\%  & All & 10\% & 5\% & All & 10\% & 5\% \\
\midrule
SPIN$^\dagger$ & - & 130.0 & 130.6 & - & - & - & - & - & - \\
PM-Net$^\dagger$ & - & 124.9 & \textbf{126.4} & - & - & - & - & - & - \\
SPIN-RT &  116.1 & 127.0 & 130.5 & 99.58 &113.5 & 114.5 & 66.53 & 77.71 & 76.66\\
\midrule
Ours (BMC) & 113.9 & 128.6 & 129.6 & 97.87 &113.7 &113.0 & 65.90 & 77.73& 76.35 \\
Ours (GAI) & \textbf{112.7} & \textbf{122.9} & 128.1 & \textbf{96.70} & \textbf{108.8} &  \textbf{111.9} & \textbf{64.69} & \textbf{74.04} & \textbf{74.35} \\

\bottomrule
\end{tabular}
\end{table*}

%% file: sections/5_conclusion.tex
\section{Discussion and Conclusion}
In conclusion, we revisit MSE's probabilistic interpretation and identify its ineffectiveness in imbalanced regression. We therefore propose a statistically principled loss function, Balanced MSE, for imbalanced regression. We further discuss various implementation options of \name{}, including both closed-form options and numerical options. \name{} outperforms existing methods on various uni- and multi-dimensional imbalanced regression benchmarks. 

\noindent\textbf{Future Works.}
Future works may use \name{} as a bridge to introduce more approaches that are developed on the imbalanced classification to the imbalanced regression. For example, \autoref{eq:bmc_loss} can be viewed as Softmax with temperature. Margin-based methods might be introduced to adjust the pair-wise distances as well. One may also leverage deep generative models, \eg, VAE~\cite{kingma2013auto} and GAN~\cite{goodfellow2014generative}, to better model $\pytrain$. 

\noindent\textbf{Broader Impacts.}
Our method only addresses the bias brought by imbalanced label distribution. However, there are still other types of biases in a training dataset besides the mentioned label imbalance. A regressor may still learn those biases and make predictions that produce negative social impacts even with the proposed method applied.

%% file: sections/6_ack.tex
\section{Acknowledgment}
This work is supported by NTU NAP, MOE AcRF Tier 2 (T2EP20221-0033), the National Research Foundation, Singapore under its AI Singapore Programme, and under the RIE2020 Industry Alignment Fund – Industry Collaboration Projects (IAF-ICP) Funding Initiative, as well as cash and in-kind contribution from the industry partner(s). 

%% file: sections/appendix.tex
\section{Proofs and Derivations}
\subsection{Proof for Theorem 1}
By Bayes Rule, we have:
\begin{align}\label{eq:appendix_bayes}
\pyxtrain = \pxy \cdot \pytrain / \pxtrain \\
\pyxbal = \pxy \cdot \pybal / \pxbal
\end{align}
By change of variables, we have:
\begin{align}\label{eq:appendix_change_of_variable}
\pyxtrain = \pyxbal \cdot \frac{\pytrain}{\pybal} \cdot \frac{\pxbal}{\pxtrain}
\end{align}
The evidence ratio $\frac{\pxbal}{\pxtrain}$ in \autoref{eq:appendix_change_of_variable} is unknown. To bypass the unknown ratio, we use the definition that the integral of $\pyxtrain$ over space $Y$ should be equal to 1. Using the simple fact, we have:
\begin{align}\label{eq:appendix_trick}
\pyxtrain =  \frac{\pyxtrain}{\int_{Y} \pyprimextrain \dyprime}.
\end{align}
Bring \autoref{eq:appendix_change_of_variable} into \autoref{eq:appendix_trick}, we have:
\begin{align}\label{eq:appendix_reparametrize}
\pyxtrain &=  \frac{\pyxbal \cdot \frac{\pytrain}{\pybal} \cdot \frac{\pxbal}{\pxtrain}}{\int_{Y} \pyprimexbal \cdot \frac{\pyprimetrain}{\pyprimebal} \cdot \frac{\pxbal}{\pxtrain} \dyprime}\\
&= \frac{\pyxbal \cdot \frac{\pytrain}{\pybal}}{\int_{Y} \pyprimexbal \cdot \frac{\pyprimetrain}{\pyprimebal} \dyprime}\\
&= \frac{\pyxbal \cdot \pytrain}{\int_{Y} \pyprimexbal \cdot {\pyprimetrain} \dyprime}
\end{align}

\subsection{MSE as a Special Case of Balanced MSE}
We show that MSE is a special case of Balanced MSE.
When $\pytrain$ is uniform on $Y$, 
\begin{equation}
\begin{split}
&\log\int_Y\gau(\boldsymbol{y};\ypred,\Sigmapred)\cdot\pytrain\dy \\ 
&= \log\int_Y\gau(\boldsymbol{y};\ypred,\Sigmapred)\cdot C \dy\\
&=\log\int_Y\gau(\boldsymbol{y};\ypred,\Sigmapred)\dy + \log C \\
&= \log 1+ \log C = \log C,\\
\end{split}
\end{equation}
where $C$ is some constant. Then, the Balanced MSE loss becomes $-\log \gau(\boldsymbol{y};\ypred,\Sigmapred) + \log C$ and is equivalent to the standard MSE loss.

\subsection{GAI Loss Derivation}\label{sect:Loss form for the GNI variant}
We continue our derivation from Eq 3.11. The integral of a Gaussian is trivial to solve:
\begin{align}\label{eq:gmm_norm}
\sum_{i=1}^K\phi_iS_i\int_Y\gau(\boldsymbol{y};\tilde{\boldsymbol{\mu}}_i, \tilde{\boldsymbol{\Sigma}}_i)\dy
=\sum_{i=1}^K\phi_iS_i
\end{align}
Therefore, the closed-form loss of Balanced MSE is:
\begin{equation}
\begin{split}\label{eq:gmm_nll}
L &=-\log \gau(\boldsymbol{y};\ypred,\Sigmapred) \\
&+ \log\int_Y\gau(\yprime;\ypred,\Sigmapred)\cdot\pyprimetrain\dyprime \\
&= -\log\gau(\boldsymbol{y};\ypred, \Sigmapred) + \log\sum_{i=1}^K\phi_iS_i
\end{split}
\end{equation}
Recall that $S_i$ is the norm of the product of two Gaussians. $S_i$ itself is also a Gaussian:
\begin{align}\label{eq:prod_norm}
S_i = \gau(\ypred;\boldsymbol{\mu}_i, \boldsymbol{\Sigma}_i + \Sigmapred)
\end{align}
Bring \autoref{eq:prod_norm} back to \autoref{eq:gmm_nll}, we have:
\begin{equation}
\begin{split}
L &= -\log \gau(\ytarget;\ypred,\Sigmapred) \\
&+ \log\sum_{i=1}^K\phi_i\cdot\gau(\ypred;\boldsymbol{\mu}_i, \boldsymbol{\Sigma}_i + \Sigmapred)
\end{split}
\end{equation}

\begin{figure*}
\centering
	\includegraphics[width=1.\textwidth]{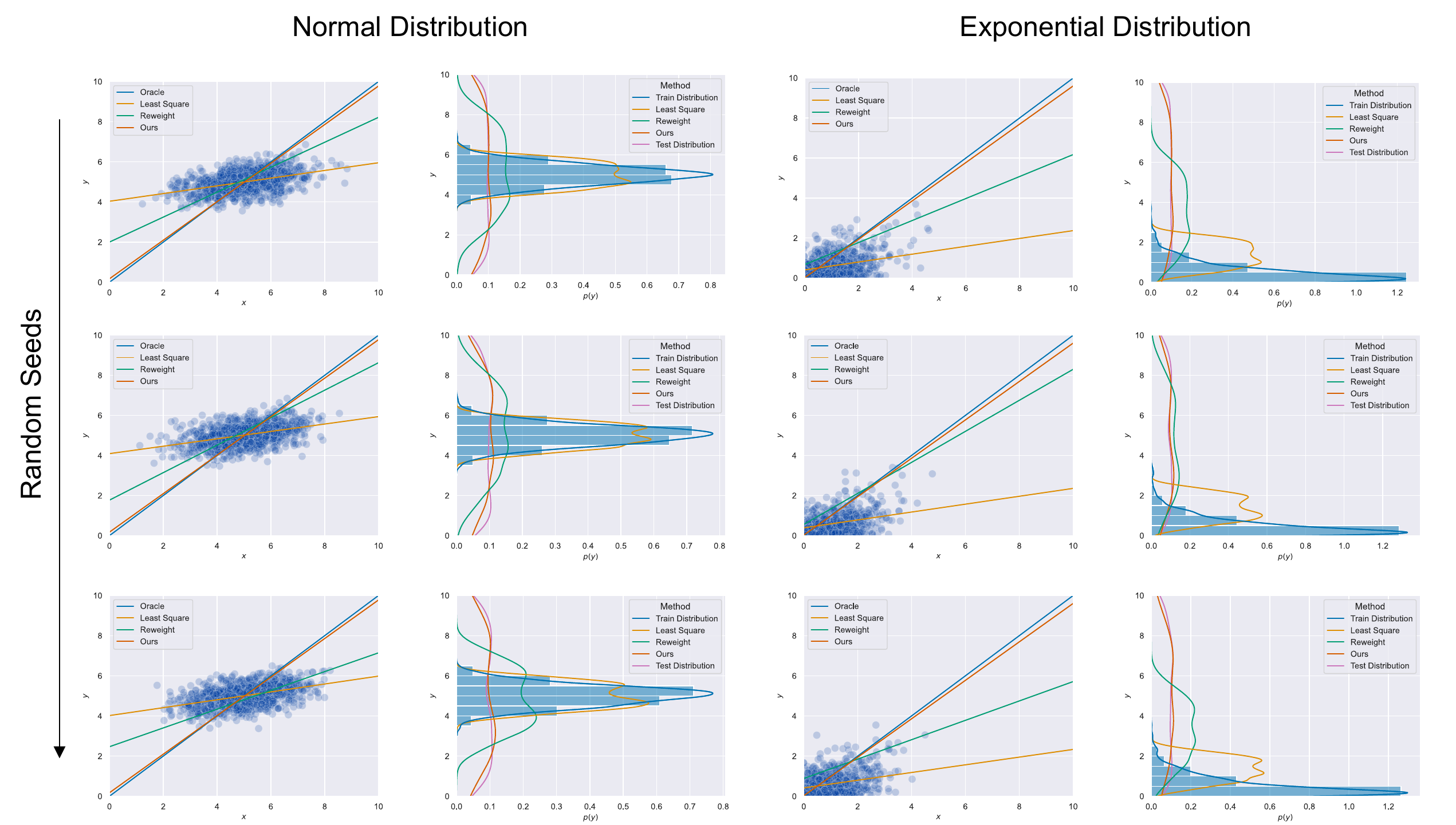}
	\caption{Synthetic benchmark on random seeds. Although the noise scale keeps the same, reweighting's performance varies drastically when different random seeds are used. In comparison, \name{} is robust to different sampled noises.} \label{fig:case_study_1d_randomseed}
\end{figure*}
\begin{figure*}
\centering
	\includegraphics[width=1.\textwidth]{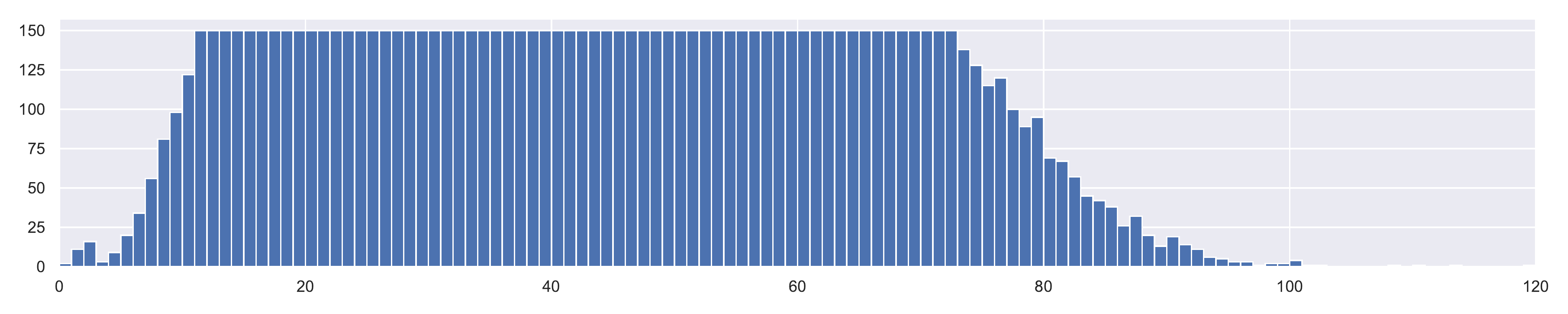}
	\caption{IMDB-WIKI-DIR test set visualization. We observe tail labels on both edges of the test distribution. Overall metrics will not sufficiently assess a model's performance on senior adults (age $> \sim$75) and children (age $< \sim$15 ).} \label{fig:testset_vis}
\end{figure*}

\section{Implementation Details}\label{sect:Implementation Details}
\subsection{Synthetic Benchmark}

\subsubsection{Dataset Construction}
For the training set, we first randomly sample 1024 labels $\boldsymbol{y}$ from a predefined label distribution $\pytrain$, \eg, a normal distribution. Then, we minus a random noise $\boldsymbol{\epsilon} \sim \gau(0,\boldsymbol{\textrm{I}})$ from the lables, to obtain the true labels $\tilde{\boldsymbol{y}}$ so that $\boldsymbol{y} = \tilde{\boldsymbol{y}} +\boldsymbol{\epsilon}$. For an invertible mapping function $f: X \to Y$, \eg, a linear function, we find its inverse function $f^{-1}$, and generate inputs $\boldsymbol{x}$ from the true labels $\tilde{\boldsymbol{y}}$ using $f^{-1}$. After that, we have:
\begin{equation}
\begin{split}
    \boldsymbol{y} &= \tilde{\boldsymbol{y}} +\boldsymbol{\epsilon} = f(\boldsymbol{x}) +\boldsymbol{\epsilon} 
\end{split}
\end{equation}
To this end, $(\boldsymbol{x}, \boldsymbol{y})$ is a standard regression dataset and $\boldsymbol{y}$ has a predefined imbalanced distribution.
We call $f$ the oracle relation and our goal is to estimate $f$ from $(\boldsymbol{x}, \boldsymbol{y})$. 

For the test set, we repeat the above procedure except that we use a uniform label distribution and do not apply the random noise.

\subsubsection{Training Details}
In training, we use a batch size 256. For one-dimensional linear regression, we train the models for 2K epochs. We use SGD optimzer with momentum 0.9. We set the learning rate to 1e-3. For non-linear regressions and two-dimensional linear regressions, we train the models for 10K epochs. We use Adam~\cite{kingma2014adam} optimizer and set the learning rate to 0.2.

\begin{figure*}
\centering
	\includegraphics[width=1.\textwidth]{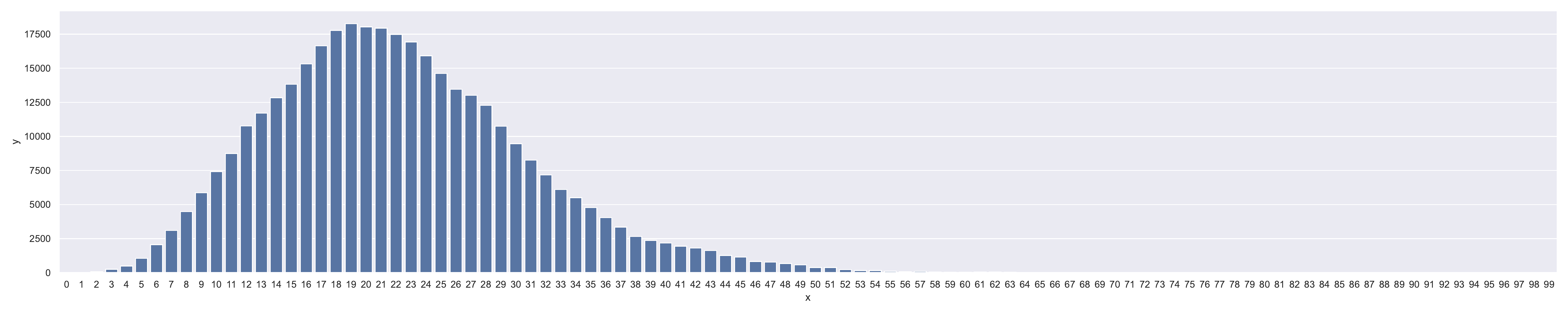}
	\caption{Visualization of the training label distribution of IHMR. The horizontal axis is 100 regions uniformly divided on the pose space according to their geodesic distance to the mean pose.} \label{fig:smpl_train_dist}
\end{figure*}

\input{tables/4_case_study_results}

\input{tables/4_ablation_on_sigma}
\input{tables/4_ablation_on_gmm}

\subsection{IMDB-WIKI-DIR}
We follow the RRT setting in~\cite{yang2021delving}. Concretely, we use ResNet-50~\cite{he2016deep} model as the backbone. We train the vanilla model for 90 epochs using Adam optimizer~\cite{kingma2014adam}. We decay the learning rate from 10\textsuperscript{-3} by 0.1 at 60-th epoch and 80-th epoch. We then freeze the backbone, re-initialize and train the last linear layer. For the retraining, we train the last linear layer for 30 epochs with a constant learning rate at 10\textsuperscript{-4}. We use a GMM with 2 components.

\subsection{NYUD2-DIR}
We follow the settings in~\cite{yang2021delving}. We use a ResNet-50-based encoder-decoder architecture proposed by~\cite{hu2019revisiting}. We train the model for 20 epochs using Adam optimizer with an initial learning rate at 10\textsuperscript{-4}. The learning rate decays by 0.1 every 5 epochs. Only direct supervision on depth is used in training. We use a GMM with 16 components.

\subsection{IHMR}
We use a pretrained SPIN~\cite{kolotouros2019learning} model as the feature extractor, and re-train the linear regressor for 20 epochs. We follow SPIN to train on the following 3D datasets: Human3.6M~\cite{ionescu2011latent}, MPI-INF-3DHP~\cite{mehta2017monocular}; 
and following 2D datasets:
LSP~\cite{johnson2010clustered};
LSP-extended~\cite{kolotouros2019learning},
MPII~\cite{andriluka20142d}, 
COCO~\cite{lin2014microsoft}. We test on 3DPW~\cite{von2018recovering}. Static fits are used to provide supervision on the 2D datasets. We use a constant learning rate at 10\textsuperscript{-4}. We use a GMM with 16 components.

\subsection{Noise Scale Learning}
We set $\sigmapred$ as a learnable variable that requires gradient, and add it into the optimizer so that $\sigmapred$ can be optimized together with model parameters. There are no additional network or architecture modifications for the noise scale learning.

\section{Experiment on random seeds}\label{sect:Synthetic benchmark on noise}
We compare least square, reweighting, and \name{} under different random seeds in the one-dimensional linear regression. A visualization of results is shown in \autoref{fig:case_study_1d_randomseed}. We observe that reweighting is sensitive to random seeds. Reweighting's performance varies drastically when random seed changes. This may attribute to the fact that reweighting signifies rare labels' noise and the zero mean noise assumption no longer holds. In comparison, \name{} is robust to different noise sampling results.

\section{Quantitative results for the synthetic benchmark}

We show the quantitative results for the synthetic benchmarks. There are three settings in the quantitative results. \textbf{Normal}: one-dimensional linear regression where the label distribution is a Normal distribution. \textbf{Exponential}: one-dimensional linear regression where the label distribution is an Exponential distribution. \textbf{MVN}: two-dimensional linear regression where the label distribution is a Multivariate Normal distribution. 

Different extents of distribution skewness are studied as well. The results show that \textbf{1)} both GAI and BMC significantly outperforms Vanilla (\ie, least square) and Reweighting, particularly when the skewness is high; \textbf{2)} the numerical implementation BMC shows comparable performance to the closed-form implementation GAI; \textbf{3)} using learned noise scale achieves a comparable performance to using the true noise scale.

\section{IMDB-WIKI-DIR test set visualization}
We visualize the label distribution of IMDB-WIKI-DIR's test set in \autoref{fig:testset_vis}.

\section{Ablations}
\subsection{Effect of the noise scale}
We study the effect of $\sigmapred$ on IMDB-WIKI-DIR, by fixing $\sigmapred$ at different values. We use the GAI option for study. We also compare fixed $\sigmapred$ (Fix.) with jointly optimized $\sigmapred$ (Joint.). Results are shown in \autoref{tab:ablation_on_sigma}. We observe that larger $\sigmapred$ trades the performance towards tail labels. We also observe that the jointly optimized $\sigmapred$ is effective in finding the optimal trade-off point.

\subsection{Effect of number of components in GMM}
We study the number of components K in GMM on IMDB-WIKI-DIR using the GAI variant. Results are shown in \autoref{tab:ablation_on_gmm}.
We notice that the performance reaches optimal when K is larger or equal to 2. This may attribute to the fact that the training label distribution of IMDB-WIKE-DIR is relatively simple.

\section{Additional Discussions and Analysis}

\input{tables/rebuttal_compute}
\noindent \textbf{Analysis on the Computational Cost.}\label{sect:computational cost}
We compare Balanced MSE with other methods in terms of computational cost in this section. We show the train-time computational cost on IMDB-WIKI-DIR in \autoref{tab:compute}. Results are averaged on the first epoch. The overhead is negligible compared to overall cost. There is no additional computational cost during inference.

\noindent \textbf{How is \name{} connected to the Bayes-optimal prediction?}
We use $\ypred$, the mean of the predicted Gaussian, to infer the final label. Since the mean and the mode are the same for a Gaussian distribution, it is by definition that $\ypred$ estimated by \name{} is the Bayes-optimal prediction for a balanced test set: $ \ypred=\textrm{argmax}_{\boldsymbol{y}}\gau(\boldsymbol{y};\ypred, \Sigmapred)=\textrm{argmax}_{\boldsymbol{y}} \pyxbaltheta$.

\noindent \textbf{Why model the noisy prediction as an isotropic
Gaussian?}
The isotropic Gaussian noise is assumed by ordinary
least square~\cite{jain2015alternating}. More fine-grained noise correlations modeling can lead to better regression performance~\cite{jain2015alternating} but is
out of the scope of \name{}.

\noindent \textbf{Will modeling the uncertainty explicitly help imbalanced regression?} 
\name{} estimates a constant noise and degrades to MSE when no imbalance exists, \ie, the gain is from imbalance handling not from uncertainty modeling. However, sophisticated uncertainty modeling, \eg, correlated noise~\cite{jain2015alternating} and input-dependent noise~\cite{le2005heteroscedastic}, could help regression in general.

\noindent \textbf{Can we extend the analysis in \name{} to L1 \& Huber loss?}
Extending L1 \& Huber loss to balanced versions will be important future works, which can be done via Theorem 1 by replacing Gaussian in this work to Laplacian and ~\cite{meyer2021alternative} respectively.

%% file: tables/4_case_study_results.tex
\begin{table*}
\small
 \centering
 \caption{Quantitative results for the synthetic benchmark. $\dagger$: True noise scale used. For each type of distribution, we evaluate three extents of skewness: Low, Moderate, and High. Best results are bolded.}
\label{tab:case_study}
\begin{tabular}{lrrrrrrrrr}

    \toprule
     & 
    \multicolumn{3}{c}{Normal (MSE$\downarrow$)} 
    &
    \multicolumn{3}{c}{Exponential (MSE$\downarrow$)} 
    &
    \multicolumn{3}{c}{MVN (MSE$\downarrow$)} \\ 
    
    \cmidrule(lr){2-4} 
    \cmidrule(lr){5-7} 
    \cmidrule(lr){8-10} 

Method & High & Mod. & Low & High & Mod. & Low &High & Mod. & Low \\
\midrule
Vanilla & 5.521 & 3.275 & 1.936 & 18.61 &13.14 & 6.038 & 5.522 &3.809 & 2.570   \\
Reweight &1.399 & 0.336 & 0.092 & 4.676 &1.336 & 0.128 & 3.310 & 1.758 &1.001    \\
\midrule
Ours (GAI)$^\dagger$ & \textbf{0.031} & \textbf{0.001} & 0.001  & \textbf{0.001} & 0.002 & 0.004 &\textbf{0.122} &0.031 &  0.011 \\
Ours (BMC)$^\dagger$ &0.043 & 0.004 & \textbf{0.000} & 0.002 &\textbf{0.000} & \textbf{0.000} &0.126 &0.033 & 0.011  \\
Ours (GAI) &0.089 & 0.008 &0.005 & 0.130 & 0.082 & 0.023  & 0.184 & \textbf{0.021} &  \textbf{0.006} \\
Ours (BMC) & 0.141 &  0.060 &0.030 & 0.122 & 0.104 & 0.034  &0.142 & 0.025& 0.011 \\
\bottomrule
\end{tabular}
\end{table*}

%% file: tables/4_ablation_on_sigma.tex
\begin{table*}[!ht]
\small
 \centering
\caption{Ablation on the choice of noise on IMDB-WIKI-DIR.}
\label{tab:ablation_on_sigma}
\begin{tabular}{lrrrrrrrr}

    \toprule
     & 
    \multicolumn{4}{c}{bMAE$\downarrow$}
    &
    \multicolumn{4}{c}{MAE$\downarrow$} \\ 
    
    \cmidrule(lr){2-5} 
    \cmidrule(lr){6-9} 

Method & All & Many & Med. & Few  & All & Many & Med. & Few \\
\midrule
Fix. ($\sigma$ = 6) &  12.85 & 7.27 &13.26  & 29.79  &7.81 &7.20  & 12.78 & 23.78  \\
Fix. ($\sigma$ = 7) &  12.67 &  7.52  &  12.75 & 28.67 &8.00 & 7.45 & 12.32 &  23.25 \\
Fix. ($\sigma$ = 8) &  12.68 &  7.80  & 12.61  & 27.83 & 8.24 & 7.73 &12.21  & 22.94 \\
Joint. & 12.66 & 7.65 & 12.68 & 28.14 & 8.12 &  7.58 & 12.27 & 23.05\\
\bottomrule
\end{tabular}
\end{table*}

%% file: tables/4_ablation_on_gmm.tex
\begin{table*}[!ht]
\small
 \centering
\caption{Ablation on the effect of the number of components K in the GMM.}
\label{tab:ablation_on_gmm}
\begin{tabular}{lrrrrrrrr}

    \toprule
     & 
    \multicolumn{4}{c}{bMAE$\downarrow$}
    &
    \multicolumn{4}{c}{MAE$\downarrow$} \\ 
    
    \cmidrule(lr){2-5} 
    \cmidrule(lr){6-9} 

Method & All & Many & Med. & Few  & All & Many & Med. & Few \\
\midrule
K=1 & 12.72 & 7.70 & 12.94 & 28.08 & 8.18 &  7.63 &12.47 &23.17\\
K=2 & 12.66 & 7.65 & 12.68 & 28.14 & 8.12 &  7.58 &12.27 &23.05\\
K=4 & 12.67 & 7.62 & 12.68 & 28.26 & 8.09 &  7.55 &12.26 &23.03\\
K=128 & 12.66 & 7.61 & 12.87 & 28.11 & 8.09 &  7.53 &12.44 &23.18\\
\bottomrule
\end{tabular}
\end{table*}

%% file: tables/rebuttal_compute.tex
\begin{table}[!ht]
\small
\centering 
\caption{Computational cost comparison.}
\label{tab:compute}
\begin{tabular}{lccr}

\toprule
{} & Time (s/iter) & Memory & Remark \\
\midrule
RRT  &  0.29 & 6502MB & -\\
LDS & 0.29 & 6502MB & - \\
\midrule
GAI & 0.30 & 6502MB & K=2 \\
GAI & 0.30 & 6512MB & K=512 \\
BMC & 0.30 & 6504MB & B=256 \\
\bottomrule
\end{tabular}
\end{table}